\documentclass[10pt,twocolumn,letterpaper]{article}

\usepackage[pagenumbers]{cvpr}

\usepackage{graphicx}
\usepackage{amsmath}
\usepackage{amssymb}
\usepackage{booktabs}
\usepackage{array}
\usepackage{color}
\usepackage{xcolor}
\usepackage[pagebackref,breaklinks,colorlinks]{hyperref}
\usepackage{multirow}% http://ctan.org/pkg/multirow
\usepackage{multicol}
\usepackage{hhline}
\usepackage{bm}
\usepackage{makecell}
\usepackage{colortbl}
\usepackage[normalem]{ulem}
\definecolor{darkcyan}{RGB}{0,127,127}
\definecolor{darkmagenta}{RGB}{127,0,127}
\definecolor{black}{RGB}{0,0,0}

\newcommand{\tablelightgray}{\rowcolor[gray]{.95}}
\newcommand{\tabledarkgray}{\rowcolor[gray]{.80}}

\usepackage[capitalize]{cleveref}
\crefname{section}{Sec.}{Secs.}
\crefname{section}{Section}{Sections}
\crefname{table}{Table}{Tables}
\crefname{table}{Tab.}{Tabs.}

\begin{document}
\title{Generalizable Deepfake Detection with Phase-Based Motion Analysis}
\author{Ekta Prashnani\thanks{Work done when EP (now at NVIDIA) was a Ph.D. student at UC Santa Barbara.}
	\qquad 
	Michael Goebel
	\qquad 
	B. S. Manjunath\\
	Department of Electrical and Computer Engineering, University of California, Santa Barbara\\
	\fontsize{10pt}{10pt}{\tt \{ekta, mgoebel, manj\}@ece.ucsb.edu}
\vspace{-0.11in}
}

\maketitle
\begin{abstract}
    %!TEX root = ../eccv2022submission.tex

We propose PhaseForensics, a DeepFake (DF) video detection method that leverages a phase-based motion representation of facial temporal dynamics. 
Existing methods relying on temporal inconsistencies for DF detection present many advantages over the typical frame-based methods.
However, they still show limited cross-dataset generalization and robustness to common distortions.
These shortcomings are partially due to error-prone motion estimation and landmark tracking, or the susceptibility of the pixel intensity-based features to spatial distortions and the cross-dataset domain shifts.
Our key insight to overcome these issues is to leverage the \textit{temporal phase variations} in the band-pass components of the Complex Steerable Pyramid on face sub-regions.
This not only enables a robust estimate of the temporal dynamics in these regions, but is also less prone to cross-dataset variations.
Furthermore, the band-pass filters used to compute the local per-frame phase form an effective defense against the perturbations commonly seen in gradient-based adversarial attacks.
Overall, with PhaseForensics, we show improved distortion and adversarial robustness, and state-of-the-art cross-dataset generalization, with 91.2\% video-level AUC on the challenging CelebDFv2 (a recent state-of-the-art compares at 86.9\%).
\vspace{-0.1in}

\end{abstract}
%!TEX root = ../main.tex
\section{Introduction}
\label{sec:intro}
%% threat posed by hyper-realistic generative models
High-quality deep generative models for faces~\cite{Kar21,Kar19}, and their seamless public access~\cite{zapp,refaceapp}), has enabled areas like art or communication.
However, they also pose a societal threat if the deepfakes generated by such models are used to propagate misinformation~\cite{Vac20,Dia21}.
This has led to the active pursuit of designing deepfake (DF) detection methods~\cite{Yu21,Cas21,Sud21}.
Broadly, DF detection methods either rely on the spatial per-frame artifacts left by DF generators (such as warping~\cite{Li18} or upsampling~\cite{Liu21}) or the anomalies in facial dynamics across frames of a DF video (such as in lip movements~\cite{Hal21}).
Leveraging the inconsistencies in the facial temporal dynamics for DF detection is more promising of the two directions, since it is harder to synthesize realistic facial motion.

\begin{figure}[t!]
	\centering
	\includegraphics[width=1.0\linewidth, trim=0cm 0cm 0cm 0cm, clip=true]{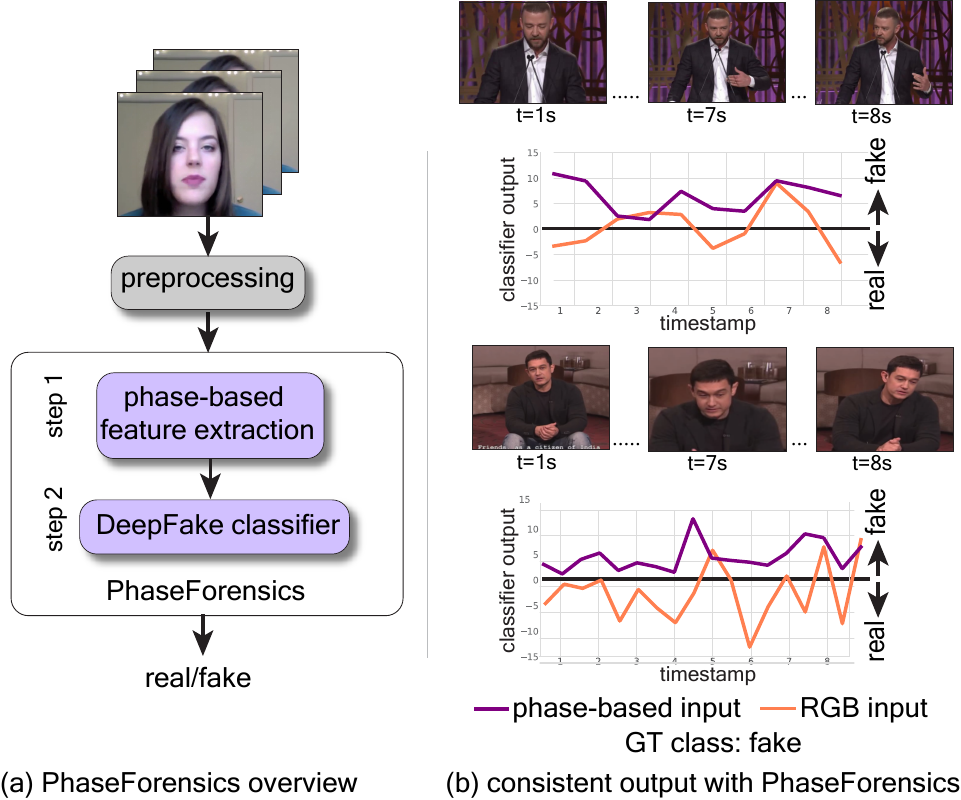}
	\vspace{-0.1in}
	\caption{In this work, we propose PhaseForensics, a novel a phase-based approach for DeepFake (DF) detection (a).
	We use spatio-temporally filtered phase from frequency sub-bands of frames to train the DF detector.
	When compared to directly using the RGB data, phase-based DF detection decreases per-frame prediction variability despite appearance changes as shown in (b). We further show this improved robustness by superior performance on unseen test data, and spatial/adversarial distortions (Tab.~\ref{tab:main_results}).}
	\vspace{-0.1in}
	\label{fig:intro}  
\end{figure}
We contribute to this promising trend with our proposed method, \textit{PhaseForensics} (Fig.~\ref{fig:intro}), that relies on temporal phase changes to estimate a noise-robust, domain-invariant, representation of the facial dynamics for DF detection.
Distinct from the existing phase-based DF detection methods~\cite{Liu21} -- that only rely on \textit{spatial} anomalies in the per-frame phase spectrum -- we use the \textit{temporal} phase changes in frequency sub-bands across frames to explicitly leverage \textit{motion}-related information.
This allows us to learn the DF classifier from the high-level facial temporal dynamics instead of the per-frame artifacts.

Traditionally, temporal methods for DF detection either use hand-coded features such as estimating motion vectors~\cite{Ame19} or landmark trajectories~\cite{Sun21}, or learn temporal semantics in an end-to-end manner~\cite{Hal21}.
These techniques can be error-prone due to factors such as dependence on estimation accuracies, or tracking errors.
Instead, we adopt an Eulerian approach to estimating motion-based features~\cite{Fle90}.
Specifically, we use the temporal phase changes across video frames in the frequency sub-bands to capture the motion field~\cite{Gau02,Wad13,Fle90}.
This circumvents the need for error-prone tracking/optical-flow estimation or solely relying on the trained model to learn motion-relevant features from RGB-domain inputs~\cite{Hal21}, but still provides an indication of the amount of motion in face sub-regions.
Using this link between phase and motion, we isolate the facial temporal dynamics with learnable spatio-temporal filters applied to the phase of the coefficients of a complex steerable pyramid (CSP) of the frames~\cite{Wad13}.
Then, we adopt a standard DF classifier pipeline with feature extraction and sequence modeling~\cite{Hal21}.

Operating in the phase domain has another crucial advantage: it affords increased robustness to appearance changes (e.g., contrast or scale changes)~\cite{Gau02,Fle90}.
Consequently, we observe improved robustness to spatial distortions associated with color, noise, and compression artifacts, as well as state-of-the-art cross-dataset generalization (e.g., with of 91.2\% in terms of AUC on Celeb-DFv2 dataset~\cite{Li20}).
In Fig.~\ref{fig:intro}b, we demonstrate this, in comparison to LipForensics~\cite{Hal21}, (which uses RGB-domain input) by plotting the classifier output for sub-clips of videos.
PhaseForensics output shows fewer oscillations as the video progresses and a consistently accurate predicted class for the video, despite the varying scale of the face and transient factors such as head pose or lighting changes.

An under-explored aspect of the performance analyses of DF detectors is evaluating their adversarial robustness. 
Recent studies reveal that existing DF detectors tend to be vulnerable to adversarial attacks, limiting their applicability~\cite{Nee21,Hus21}.
With PhaseForensics, we observe a higher adversarial robustness compared to existing temporal state-of-the-art methods such as LipForensics~\cite{Hal21}, for black-box attacks.
Previous works have shown that adversarial attacks typically target the high-frequency components of image inputs~\cite{Wang20B}.
While estimating phase variations to capture the motion field, the input features estimated for PhaseForensics are obtained from the band-pass components, discarding the higher frequencies. 
These input features enable our deep learning model to learn from lower frequency components, yielding adversarial robustness by design.

Overall, with PhaseForensics, we achieve state-of-the-art cross-dataset generalization, improved or -- in some cases -- on-par robustness to spatial distortions compared to existing methods, and improved adversarial robustness to black-box attacks (Fig~\ref{fig:intro}).
We analyze the design choices of our proposed method through experiments, to clearly demonstrate the advantage of each of the components.

%!TEX root = ../main.tex
\section{Related work}
\label{sec:related_work}

\begin{figure*}[t!]
	\centering
	\vspace{0.1in}
	\includegraphics[width=1.0\linewidth, trim=0cm 0cm 0cm 0cm, clip=true]{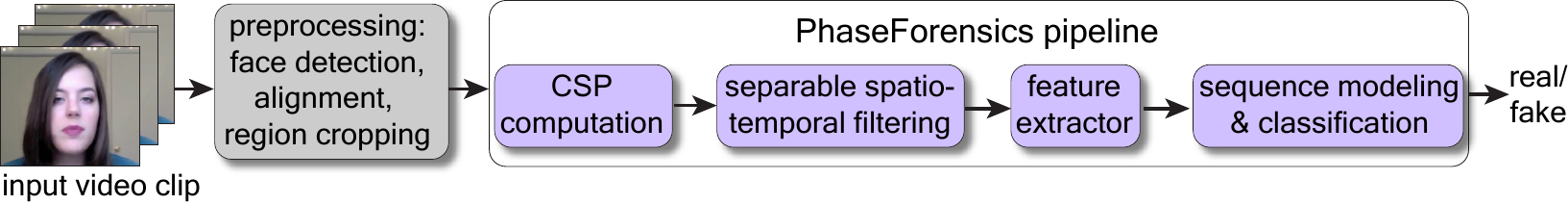}
	\vspace{-0.1in}
	\caption{\textbf{PhaseForensics overview.} For a given video clip of a facial sub-region, we compute the per-frame complex steerable pyramid (CSP) decomposition, followed by spatial and temporal filtering of the phase in frequency sub-bands to isolate motion cues suitable for deepfake detection. The spatio-temporally filtered phase is then passed to a ResNet-18 feature extractor followed by sequence modeling and classification.}
	\label{fig:method_overview}
	\vspace{-0.15in}
\end{figure*}
We now discuss the existing notable works for DF detection, and refer the readers to comprehensive reviews on the topic for a more in-depth discussion~\cite{Yu21,Cas21}.

\vspace{0.05in}
\noindent\textbf{Frame-based methods.} The sub-field of frame-based DF detection has witnessed active progress over the years, with early methods performing per-frame feature-extraction using pretrained networks~\cite{Ros19} to more sophisticated frame-based approaches utilizing attention, for example: \cite{zhao2021multi} perform fine-grained per-frame classification using the learnt multi-attentional maps with soft-attention dropout and a regional independence loss, \cite{Wan21} propose an image-based method that uses multi-scale transformers to process RGB image, followed by fusion of processed features with the DCT domain. 
The loss used for training is a combination of cross-entropy, segmentation, and contrastive loss.
While frame-based methods have been popular, dependence on single-frame DF generation artifacts can be prone to inaccuracies.
This is because of the rising sophistication of DF generators~\cite{Kar21,Kar19}, and also because the imprints left by such generators can easily be lost by simple architecture changes~\cite{Chan21}.
This can especially impact methods that rely on signals such as generative model footprint~\cite{Liu21}, face warping artifacts~\cite{Li18}, or blending boundaries for classifying real vs. fake faces~\cite{LiL20B}.
Therefore, several other methods have attempted to improve detection performance and generalization by leveraging additional features.
Such features have included Discrete Cosine Transform (DCT) coefficients~\cite{Qia20}, mid-level features~\cite{Afc18}, and Laplacian of Gaussian filtered inputs~\cite{Mas20}.

\vspace{0.05in}
\noindent\textbf{Methods relying on biological signals.} Many methods rely on detecting inconsistencies in facial landmarks between real and deepfake videos~\cite{Sun21,Dem21}. These methods leverage existing, pretrained face landmark detection models, to extract a condensed feature to be passed for classification. 
While these methods may offer a degree of explainability not obvious in more complicated neural networks, the performance results generally lag behind the end-to-end trainable models.
Related to these are the methods that rely on biological signals -- with the aim to track inconsistencies in heart-rate~\cite{Her20,Das21,Cif20A,Cif20B} or blink/gaze patterns~\cite{Jun20,Dem21,Cif20A}).
With these approaches, there is a susceptibility towards errors since such methods depend on reliable estimation of such signals -- which can be lost for low-quality videos.

\vspace{0.05in}
\noindent \textbf{Methods leveraging phase/frequency information.} An important sub-class of DF detection methods have considered leveraging frequency and phase-related cues. 
Specifically, these include phase or frequency spectrum imprint of generative models~\cite{Liu21,Fra20,LiJ21,Luo21}, per-frame Laplacian of Gaussians~\cite{Mai21}, and DCT coefficients~\cite{Qia20}. 
All these methods use the frequency or phase domain to estimate frame-level features to inform DF detection.
In contrast, our novelty lies in effectively leveraging \textit{temporal} phase variations to estimate a domain-invariant representation of local \textit{motion} in face regions.
Therefore, we leverage a fundamentally different aspect for DF detection compared to existing methods that use phase/frequency.

\vspace{0.05in}
\noindent \textbf{Methods that perform temporal processing.} In order to overcome the potential limitations of frame-based approaches, a recent trend has been to perform sequence modeling -- to make the DF detector dependent on temporal signals.
Improved performance by using per-frame backbone models has therefore been observed, with output features being fed into Recurrent Neural Networks (RNNs)~\cite{Gue18,Sab19,He16,Hua17} or Long-Short Term Memory (LSTM)~\cite{Cho17,Chi20} networks, and trained end-to-end~\cite{Gue18,Sab19,He16,Hua17}. 
Besides these, there have been several promising recent developments.
\cite{Mas20} present a two-branch RNN-based spatio-temporal deepfake detector, utilizing Laplacian of Gaussian to amplify medium and high-frequency artifacts, while the second branch uses RGB data. 
\cite{Sun21} model the temporal behavior of landmarks is fed to a two-stream RNN for classification. 
\cite{Hal21} propose LipForensics: a multi-scale temporal CNN is used to classify deepfakes based on temporal features extracted from lip region using a ResNet-18 and 3DCNN layers. An important step adopted here is the domain-specific pretraining on lip-reading task. The learned high-level lip semantics afford high generalizability to novel datasets and robustness to perturbation.
\cite{Lu21} learn spatial and temporal attention maps used to modulate shallow and mid-level features with the aim of capturing long-spatial-distance dependencies and anomalies in the coordination of facial features.
Despite recent advances, we observe that the generalization and robustness of temporal DF detectors can be improved. 
To achieve this with PhaseForensics, we move away from depending on pixel-intensity-based features, and instead rely temporal on phase variations robustly isolate the facial dynamics for DF detection. 
We now discuss the details of our approach.

%!TEX root = ../main.tex

\section{PhaseForensics: Phase-based DF detection}
\label{sec:method}

For any given input video, we first apply a standard pre-processing pipeline to detect the face region in the input video, stabilize it, and crop out the relevant portion.
After this, we apply the two-step PhaseForensics pipeline (Fig.~\ref{fig:intro}a).
First, we estimate the local per-frame phase from spatial sub-bands (using the complex steerable pyramid~\cite{Por00}), which we then pass to a learnable spatio-temporal filter to isolate the relevant facial dynamics from phase to inform DF detection.
Second, we use these spatio-temporally filtered phase-based features as input to train a standard DF detection pipeline comprising of a feature extractor and a multi-scale temporal convolutional network that is known to effectively perform sequence modeling~\cite{Mar20,Hal21,Lin20}.
We now elaborate upon each of these steps, along with a discussion of alternate design choices (see Sec.~\ref{sec:results} for analysis).

\subsection{Spatio-temporal feature estimation}
\label{subsec:phase_proc}
For an image, $I(x,y)$, global translations along  $x$ and $y$ directions, such as $I(x+\Delta x,y+\Delta y)$, directly relate to the phase changes in the Fourier coefficients of $I(x,y)$. 
Specifically, for all frequency components $(\omega _x,\omega _y)$, the phase change due to the translation would correspond to $(\omega _x \Delta x,\omega _y \Delta y)$ as per the Fourier Shift Theorem.
Given their infinite spatial support, phase changes in the Fourier basis functions directly capture \textit{global} translation in $I(x,y)$.
Intuitively, this idea can be extended to \textit{local} spatial shifts: looking at the phase changes in an image decomposition constituted by finite-support quadrature filter pairs -- such as with complex wavelets, or, its polar-separable version, the complex steerable pyramid~\cite{Por00} -- provide an indication of the local shift in image content.
In a typical video, $V(t,x,y)$, (e.g, that of a talking face), motion is spatially localized, and can be captured with such local phase changes.
This motion may differ across multiple scales, and may not be effectively captured by a single motion estimation filter.
To represent these multiple granularities of motion, we utilize the complex steerable pyramid (CSP) decomposition~\cite{Por00} and compute the phase of the complex coefficients obtained from the scaled and oriented bandpass components of the per-frame CSP.
Temporal changes in the phase of these components represents local motion in the video at the different orientations and scales. 
The low-pass and high-pass residual components of CSP, being real-valued, do not contain phase information.
We therefore only consider the bandpass components of the CSP in PhaseForensics.
Moreover, since these residual components directly relate to image intensities, they are prone to spatial perturbations and domain-specific cues.
We analyze this in Sec.~\ref{sec:results}, where we compare the result of using CSP residuals for training as well.
As an aside, we note that CSP is also a popular and effective choice for estimating local motion for other video-related tasks such as motion magnification~\cite{Wad13}, motion interpolation~\cite{Mey15}, and motion transfer~\cite{Pra17}.

\begin{figure}[t!]
	\centering
	\includegraphics[width=1.0\linewidth, trim=0cm 0cm 0cm 0cm, clip=true]{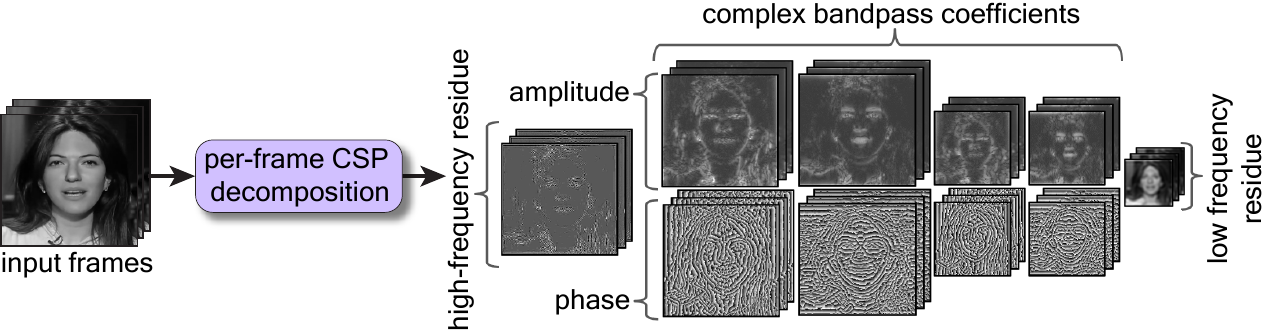}
	\vspace{-0.2in}
	\caption{Here we visualize the per-frame output of a $4$-scale, $2$-orientation complex steerable pyramid decomposition (zoom in for details). The phase of the bandpass filtered is utilized for next steps of the PhaseForensics pipeline (Fig.~\ref{fig:method_overview}).}
	\label{fig:phase}
	\vspace{-0.1in}

\end{figure}

For the input video frame at time instance $t$, $V(t,x,y)$, the complex-valued coefficient, $R_{\omega, \theta}(t,x,y)$, obtained after filtering with the spatial bandpass filter, $\Psi_{\omega,\theta}(x,y)$~\cite{Por00}, at scale $\omega$ and orientation $\theta$ is computed as:
\vspace{-0.05in}
\begin{align}
    \begin{split}\label{eq:csp_response}
        R_{\omega, \theta}(t,x,y) = {}& V(t,x,y) \ast \Psi_{\omega,\theta}(x,y)\\
                        = {}& A_{\omega, \theta}(t,x,y) (C_{\omega, \theta}(t,x,y) \\
        {}&+ iS_{\omega, \theta}(t,x,y)).
    \end{split}
\end{align}
Here, $A_{\omega, \theta}(\cdot)$ is the amplitude of the complex response, and $A_{t,\omega, \theta}(\cdot) C_{t,\omega, \theta}(\cdot)$, $A_{t,\omega, \theta}(\cdot) S_{t,\omega, \theta}(\cdot)$ denote the responses to the quadrature filter pairs, with phase $\phi_{t,\omega, \theta}(x,y)=\text{arctan}(S_{\omega, \theta}(t,x,y)/C_{\omega, \theta}(t,x,y))$.
The phase estimates from all scales and orientations of the CSP are concatenated along a new dimension, which we term $c$, to yield a per-frame tensor at each instance $t$ which we denote as $\Phi(t,x,y,c)$.

Given $\Phi(t,x,y,c)$ for all the video frames, we now want to isolate phase variations across time that are relevant to DF detection.
Traditionally, the motion of relevant objects is isolated by a hard-coded temporal filter $\Phi(t,x,y,c)$~\cite{Oh18}.
However, harcoding a temporal filter design can be suboptimal for DF detection: for example, when capturing motion of lip region, it is hard to guarantee that a hardcoded temporal filter would capture lip movements at different talking speeds.
Instead, we allow our DF detector to guide this process of isolating the temporal phase changes relevant to DF detection through an end-to-end learning process.
Therefore, we apply a learnable temporal filtering operation $f_t$ (with parameters $\theta_f$) to $\Phi(t,x,y,c)$.
Before temporal filtering, we perform a spatial filtering operation (again, learnable -- with parameters $\theta_s$) $s_{x,y}$ on $\Phi(t,x,y,c)$, to overcome any spurious spatial artifacts and improve the signal-to-noise ratio of $\Phi(t,x,y,c)$ (previous work do this with a hard-coded filter~\cite{Wad13}).
A noteworthy issue with $\Phi(t,x,y,c)$ is the ambiguity around large motions since phase shifts beyond $2\pi$ are ill-defined.
One way to resolve this to some extent is explored in motion interpolation works~\cite{Mey15} -- where phase information of different scales are used to resolve the ambiguity for a given scale.
With our proposed construction of $\Phi(t,x,y,c)$, our deep learning model has the flexibility to also perform this operation.
This is because the scale sub-bands in $\Phi(t,x,y,c)$ are stacked along the channel dimension -- allowing for across-scale interactions in a typical conv layer.
Specifically, in the implementation of the 3D separable convolution, the receptive field of $f_t$ and $s_t$ along the channel dimension ensures this interaction along the various sub-bands, which can help resolve the ambiguity.
In summary,  the post-processing operations on $\Phi(t,x,y,c)$ are implemented as a learnable, separable, $3D$ convolution (as done in~\cite{Xie18}), to yield,
\begin{equation}
    \Phi^p(t,x,y,c) = f_t(s_{x,y}(\Phi(t,x,y,c))),
    \label{eq:phase_postproc}
\end{equation}
which we use for our DF detection (super-script $p$ indicates processed output).

\vspace{0.08in}
\noindent \textbf{Alternatives to CSP.}
An alternative to obtain phase information is discrete Fourier transform (DFT): as discussed earlier, the infinite spatial support of the DFT basis makes it impossible to use its coefficients to estimate of \textit{local} motion, while the finite spatial support of the CSP basis allows for this. 
Other sub-band decompositions can include wavelets or Laplacian of Gaussians~\cite{Mas20}, both of which, being real-valued, do not provide local phase information.
This makes CSP the best choice amongst alternatives. 

\subsection{DeepFake detection training}
\label{subsec:train_deets}
\noindent \textbf{Facial region selection.} Before training with $\Phi^p(t,x,y,c)$ for DF detection, we want to understand which facial regions can benefit the most from such an approach.
This is because, for regions that contain very little or no motion, variations in $\Phi^p(t,x,y,c)$ are meaningless~\cite{Wad13}, and can adversely affect the training process.
Since the lip region of the face provides the largest motion cues, for our main experiments we only focus on the lip region of the face to compute $\Phi^p(t,x,y,c)$, similar to a previous work, LipForensics~\cite{Hal21}.
In Sec.~\ref{subsec:ablation}, we apply PhaseForensics to the eye region and in the supplementary document we also show the experiments with training on full face, and compare it to our experiments with the lip region.
Our process of sub-region extraction and alignment follow a standard pipeline -- with face detection, landmark detection and alignment, followed by cropping (details in the supplementary document). 
We note that the alignment process is global, and does not distort the facial features. 

\noindent \textbf{Architecture.} For a given video clip, the spatio-temporally processed phase, $\Phi^p(t,x,y,c)$ (Eq.~\ref{eq:phase_postproc}) is passed to the next stage of the DF detection pipeline.
We adopt a standard two-step approach for the rest of our DF detection pipeline: we first compute per-frame feature embeddings, followed by sequence modeling to learn the temporal behavior. 
We choose to use the ResNet-18 architecture (with modified input channels to match the maximum value of $\Phi^p(t,x,y,c)$) as our feature extractor -- denoting this function as $r_{x,y}$, with learnable parameters $\theta_r$.
For temporal modeling, we choose a multi-scale temporal convolutional network (MSTCN)~\cite{Bai18}, given their remarkable performance gain over typically-used LSTMs for a variety of tasks~\cite{Mar20,Hal21,Lin20} and due to their flexible design with varying temporal receptive field and lightweight construction.
The output of MSTCN is passed through a linear layer to yield the final prediction.
We denote the function represented by MSTCN + linear classifier as $m_{t,x,y}$, with learnable parameters $\theta_m$.
The output logit from PhaseForensics pipeline for a given input video clip $V(t,x,y)$, is then given by $\widehat y = m_t(r_{x,y}( \Phi^p(t,x,y,c)))$.
We train PhaseForensics in an end-to-end manner to optimize all parameters, $\theta_f,\theta_s,\theta_r,\theta_m$ by minimizing the average binary cross-entropy loss across training samples $\frac{1}{N}\sum^N_i=\mathcal{L}_{BCE}(\widehat y^i, y^i; \theta_f,\theta_s,\theta_r,\theta_m)$, where $i$ indexes the training samples and $N$ is the training dataset size.

\noindent \textbf{Pretraining and hyperparameters.} Before training PhaseForensics for DF detection, we perform domain-specific pretraining on the lip and eye regions, with $\Phi^p(t,x,y,c)$ (Eq.~\ref{eq:phase_postproc}) as input.
We pretrain the architecture discussed above on tasks relevant to these facial sub-regions, since these are shown to significantly improve the generalization of the DF detector~\cite{Hal21}.
We note that many competing methods adopt unique training steps best-suited for their specific goals, such as pretraining on lipreading~\cite{Hal21}, ImageNet pretraining~\cite{Ros19,Mas20}, or using custom dataset~\cite{Li20}.
In our evaluation of competing methods, we regard the author-prescribed training steps as the best practice -- without taking away any of the key steps.
This is also commonly done in performance comparisons presented by existing state-of-the-art DF detection methods~\cite{Hal21,Zhe21}.
Similar to LipForensics~\cite{Hal21} (a state-of-the-art method leveraging anomaly detection), training PhaseForensics on, say, lips, involves first learning the distribution of natural temporal dynamics of lips using the Lip Reading in the Wild (LRW) dataset~\cite{Chu16}, for $10$ epochs, with a cross-entropy loss (similar to~\cite{Hal21}), and then identifying deviations from it during the training for DF detection.
More formally, there are two steps: 1) learning the natural lip movements by training for lipreading, 2) learning to detect anomalies in the lip movements of deepfakes by training on DF dataset.
In the supplementary document, we clearly motivate this two-step approach by demonstrating the drop in performance observed when LRW pretraining is not performed.
Similarly for the eye region, we pretrain for gaze prediction on the EVE dataset~\cite{Par20} for $50,000$ iterations, with crops from face images showing both the eyes, and an angular loss~\cite{Par20}, that measures the error between predicted and true gaze.
For computing $\Phi(t,x,y,c)$, CSP decomposition is composed of $4$ spatial scales and $2$ orientations for lip sub-images, and $3$ spatial scales and $4$ orientations for eye region.
%We experiment with alternative CSP decompositions in the supplementary.
For all the training steps, Adam optimizer with a learning rate of $2e^{-4}$ and a batch size of $32$ is used. 
The DF detection training is stopped when the validation loss does not show any improvement for $20,000$ training iterations.
More training details can be found in the supplementary material.

%!TEX root = ../main.tex
\section{Results}
\label{sec:results}
\label{subsec:howegneral}
\begin{figure}[t!]
	\centering
	\includegraphics[width=\columnwidth, trim=0cm 0cm 0cm 0cm, clip=true]{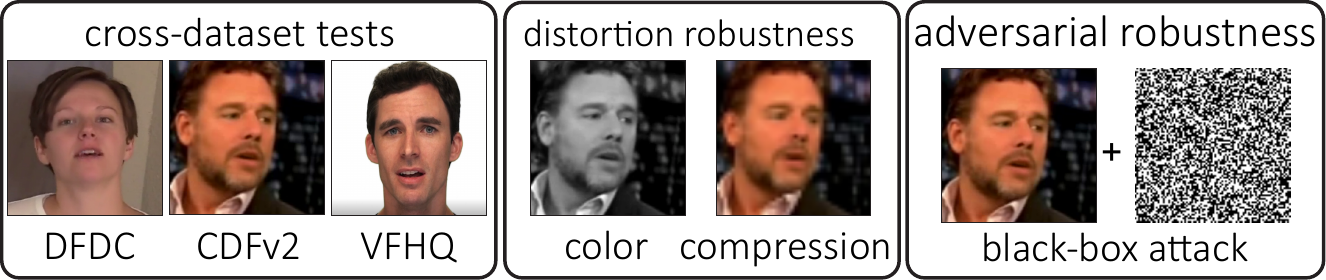}
	\vspace{-0.15in}
	\caption{\textbf{Overview of performance evaluations.} With PhaseForensics, we demonstrate state-of-the-art cross-dataset generalizability with experiments on CDFv2, DFDC and VFHQ. We also assess the robustness to spatial distortions and adversarial perturbations -- for which phase-based processing is very beneficial.}
	\vspace{-0.15in}
	\label{fig:exps_overview}
\end{figure}

\begin{table*}[t!]
    \centering
    \vspace{0.05in}
	\begin{minipage}{\columnwidth}
		\centering
		\resizebox{0.98\columnwidth}{!}{\centering

\begin{tabular}[b]{ m{2.9cm} | m{1cm} | m{1cm} | m{1.1cm}}

\Xhline{2\arrayrulewidth}
%\textbf{METHOD} &  FF++~\cite{Ros19} & FSh~\cite{LiL20A}  & DFor~\cite{Jia20} & DFDC~\cite{Dol20} & VFHQ~\cite{Fox21} &
\tabledarkgray
\textbf{METHOD} &  DFDC& VFHQ& CDFv2\\
\hline
Xception~\cite{Ros19}  & 70.9 & 70.1 & 73.7 \\
\tablelightgray
Multi-Attention~\cite{zhao2021multi} & 63.0 & 55.0 & 68.0 \\
PatchForensics~\cite{Cha20} & 65.6 & - & 69.6\\
\tablelightgray
Face X-ray~\cite{LiL20B} & 65.5 & - & 79.5\\
CNN GRU~\cite{Sab19} & 68.9 & 66.0 & 69.8\\
\tablelightgray
Two-branch~\cite{Mas20} & - & - & 76.7\\
%Multi-task~\cite{Ngu19B} & 68.1 & &75.7 &\\
DSP FWA~\cite{Li18} & 67.3 & 69.0 & 69.5 \\
%TAR~\cite{Lee21} & & &\\
%LRNet~\cite{Sun21} & & &\\
\tablelightgray
LipForensics~\cite{Hal21} & 73.5 & 90.2 & 82.4\\
FTCN~\cite{Zhe21} & 74.0 & 84.8 & 86.9\\
\hline
\tabledarkgray
PhaseForensics & \textbf{78.2} & \textbf{94.2} & \textbf{91.2} \\
\Xhline{2\arrayrulewidth}
\end{tabular}
}
		\vspace{0.07in}
		\resizebox{0.5\columnwidth}{!}{(a) cross-dataset generalization}
	\end{minipage}%
	\hfill
	\begin{minipage}{\columnwidth}
		\centering
		\resizebox{0.98\columnwidth}{!}{\centering

\begin{tabular}[b]{ m{2.9cm} | m{1cm} | m{0.9cm} | m{0.9cm}} 

\Xhline{2\arrayrulewidth}
%\textbf{METHOD} &  FF++~\cite{Ros19} & FSh~\cite{LiL20A}  & DFor~\cite{Jia20} & DFDC~\cite{Dol20} & VFHQ~\cite{Fox21} &
\tabledarkgray
\textbf{METHOD} &  FF++& DFor & FSh\\
\hline
Xception~\cite{Ros19}  & 99.8 & 84.5& 72.0 \\
\tablelightgray
Multi-Attention~\cite{zhao2021multi} & 89.8 & 72.3 & 60.2 \\
PatchForensics~\cite{Cha20} & \textbf{99.9} & 81.8& 57.8 \\
\tablelightgray
Face X-ray~\cite{LiL20B} & 99.8 & 86.8& 92.8 \\
CNN GRU~\cite{Sab19} & \textbf{99.9} & 74.1& 80.8 \\
\tablelightgray
Two-branch~\cite{Mas20} & 99.1 & - & -  \\
%Multi-task~\cite{Ngu19B} & & 77.7& 66.0& \\
%\tablelightgray
DSP FWA~\cite{Li18} & 57.5 & 50.2& 65.5 \\

\tablelightgray

LipForensics~\cite{Hal21} & \textbf{99.9} & 97.6 & 97.1 \\
FTCN~\cite{Zhe21} & 99.7 & \textbf{98.8} & \textbf{98.8} \\
\hline
\tabledarkgray
PhaseForensics & 99.5 & 97.4& 97.4 \\
\Xhline{2\arrayrulewidth}
\end{tabular}
}
		\vspace{0.07in}
		\resizebox{0.57\columnwidth}{!}{(b) cross-manipulation generalization}
	\end{minipage}
	\vspace{-0.1in}
	\caption{Here we show the video-level AUC (\%) for all models trained with FaceForensics++ (FF++)~\cite{Ros19}. Some numbers are reported from existing benchmarks~\cite{Hal21,Li20}, while a `-' is stated when the metric is not available (missing code / trained model).
	(a) shows the cross-dataset generalization on the test subsets of DFDC~\cite{Dol20}, VFHQ~\cite{Fox21}, and CDFv2~\cite{Li20} datasets, and (b) evaluates the trained models on FF++ test videos, and also with different DF generators used to manipulate original FF++videos.
	PhaseForensics achieves state-of-the-art results for cross-dataset generalization, as compared to existing methods.
	In (b), PhaseForensics ranks amongst the top-3 methods when tested on novel (not seen in training) DF generation methods applied to the original FF++ test videos (FSh~\cite{LiL20A} and DFor~\cite{Jia20}), with a performance drop of only 1.4\% AUC compared to state-of-the-art.
    Unlike PhaseForensics, most methods achieve high accuracy on the within-domain FF++ test set, while generalizing poorly to new datasets.}
	\vspace{-0.1in}    
	\label{tab:main_results}
\end{table*}

We now compare PhaseForensics to existing popular and state-of-the-art DF detection methods (Fig.~\ref{fig:exps_overview} shows an overview of our evaluation steps).
In our evaluations we consider the classic methods such as the Xception baseline~\cite{Ros19}; recent popular approaches such as PatchForensics~\cite{Cha20} (truncated Xception classifier trained on aligned faces, with result averaged over patches~\cite{Hal21}), Multi-Attention~\cite{zhao2021multi}, CNN-GRU~\cite{Sab19} (DenseNet-161~\cite{Hua17} trained with GRU~\cite{Cho14}), Face X-ray~\cite{Li20} (from~\cite{Hal21}, trained with blended images and fake samples), DSP-FWA~\cite{Li18}; and also state-of-the-art methods LipForensics~\cite{Hal21} and FTCN~\cite{Zhe21}.
Links to the source code and pretrained models provided by the authors of each of these papers are available in the supplementary materials.
For PhaseForensics, we train the model on only the lip region (Sec.~\ref{subsec:train_deets} -- since we obtain better the performance for this sub-region; discussed further in Sec.~\ref{subsec:ablation}).
LipForensics is an important baseline in our evaluations, since it also operates on the lip regions~\cite{Hal21}.
In our ablation studies, we compare against LipForensics to demonstrate the advantage of using phase over pixel intensities. 
Consistent with recent approaches~\cite{Hal21,Ros19}, we report the \textit{video}-level Area Under ROC Curve (AUC) metric (the result is averaged over frames for frame-based methods).

\vspace{0.05in}
\subsection{Evaluating the generalizability of DF detectors}
\noindent\textbf{Training dataset.} PhaseForensics and all other methods evaluated here are trained on FaceForensics++ (FF++) training set~\cite{Ros19}.
FF++ comprises of a total of $1000$ unmanipulated videos, and corresponding manipulated videos with $4$ DF generation methods -- $2$ each for face-swapping (DeepFakes~\cite{df}, FaceSwap~\cite{fs}) and face re-enactment (Face2Face~\cite{Thi16}, NeuralTextures~\cite{Thi19}) DF generation approaches.
We adopt the train / val / test splits specified by the dataset and use the trained model from FF++ training for all experiments.

\vspace{0.05in}
\noindent\textbf{Evaluation approach, datasets, and metrics.}
Recent DF detection methods tend to show near-perfect performance when evaluated on \textit{within}-domain videos (i.e., test set of FF++) -- which does not give a clear sense of real-world generalizability of such methods.
Our main aim in this section is, therefore, to thoroughly analyze the generalization to completely new datasets never seen during training.
For this (Sec.~\ref{subsec:howegneral}), we use three datasets: CelebDFv2 (CDFv2) -- containing very high-quality face-swapped deepfakes)~\cite{Li20}; DFDC test subset -- featuring face-swapping and face re-enactment deepfakes~\cite{Dol20}, and the VideoForensicsHQ dataset -- a very high-quality face re-enactment DF dataset (VFHQ)~\cite{Fox21}.
We also analyze the robustness to spatial distortions and adversarial perturbations in this cross-dataset evaluation setting (Fig.~\ref{fig:exps_overview} shows an overview of our main experiments).
Lastly, we evaluate the trained models on FF++ test videos with the same face manipulations as seen in training, and on also datasets that feature novel face manipulation methods applied to the FF++ videos, such as DeeperForensics (DFor)~\cite{Jia20} (without spatial distortions), and FaceShifter (FSh)~\cite{LiL20A}.

\vspace{0.05in}
\noindent\textbf{Cross-dataset generalization.}
Given the variety in data capture conditions and face manipulation algorithms (from both face swap and face re-enactment categories), the cross-dataset performance evaluation (Tab.~\ref{tab:main_results}) on CDFv2, VFHQ, and DFDC  allows for an in-depth assessment of the generalization capabilities of DF detectors.
PhaseForensics yields a significant improvement in cross-dataset generalization over state-of-the-art, such as LipForensics (AUC $82.4\%$ on CDFv2) ~\cite{Hal21} and FTCN (AUC $86.9\%$ on CDFv2)~\cite{Zhe21}, with a performance of $91.2\%$ on CDFv2, $94.2\%$ on VFHQ, and $78.2\%$ on DFDC.
This is due to our method's improved robustness to appearance changes (Sec.~\ref{sec:intro}), which allow for stable predictions despite the cross-dataset domain shifts.

\begin{table*}[!h]
    \centering
    \centering
\scalebox{1.0}{
\begin{tabular}{ m{2.2cm} | m{1.1cm} | m{1.2cm} | m{1.3cm} | m{1.2cm} | m{1.3cm} | m{1.2cm} | m{1.3cm}}

\Xhline{2\arrayrulewidth}
\tabledarkgray
\multicolumn{2}{c|}{\textbf{Distortion}}&\multicolumn{2}{c|}{\textbf{CNN-GRU}}&\multicolumn{2}{c|}{\textbf{LipForensics}}&\multicolumn{2}{c}{\textbf{PhaseForensics}}\\
\Xhline{2\arrayrulewidth}
type & level & \%auc $\uparrow$ & \%mape $\downarrow$ & \%auc $\uparrow$ & \%mape $\downarrow$ & \%auc $\uparrow$ & \%mape $\downarrow$ \\
\Xhline{2\arrayrulewidth}

\tablelightgray
contrast&1& 73.6& 5.4 & 74.8 & 9.2 & \textbf{89.5} & \textbf{1.9}\\
\tablelightgray
change&2& 73.5& 5.3 & 74.7 & 9.3 & \textbf{90.0} & \textbf{1.3} \\

color&1& 70.5& 1.0 & 72.8 & 11.6 & \textbf{90.9} & \textbf{0.3} \\
saturation&2& 69.4& 0.6 & 72.2 & 12.4 & \textbf{90.9} & \textbf{0.3}\\

\tablelightgray
&1& 64.8& 7.2 & 76.3 & 7.4 & \textbf{91.3} & \textbf{0.1}\\
\tablelightgray
\multirow{-2}{*}{pixelation}& 2& 62.7 &  10.2 & 74.6 &  9.5 & \textbf{84.3} & \textbf{7.6}\\

& 1 & 70.5 &  \textbf{1.0} & \textbf{74.3} & 9.8 & 72.6 & 20.0\\

\multirow{-2}{*}{compression}& 2 & 69.4 & \textbf{0.5} & \textbf{72.2} & 12.7 & 70.3 & 22.9 \\
\tablelightgray
\multicolumn{2}{c|}{black box adv. attack}& 43.4 & 37.8 & 49.5 & 39.9 & \textbf{71.1} & \textbf{22.0}\\

\Xhline{2\arrayrulewidth}

\end{tabular}
}
    \vspace{-0.1in}
    \caption{\textbf{Robustness to spatial distortions and black-box adversarial attacks.} We assess the robustness to color and compression-related spatial distortions (derived from DFor dataset) for PhaseForensics, and two competing temporal DF detection methods: CNN-GRU~\cite{Gue18}, and LipForensics~\cite{Hal21}. 
    We apply these distortions to CDFv2 and report \%AUC of each DF detector when tested on the distorted versions of the videos.
    PhaseForensics outperforms both baselines on the majority of distortion-robustness tests.
    We also report the mean absolute percentage error (MAPE) in \%AUC compared to the performance of each method on undistorted CDFv2.
    Lastly, we evaluate adversarial robustness (last row) of the three methods for a black-box attack~\cite{Hus21}. }
    \vspace{-0.1in}
    \label{tab:distortions}
\end{table*}

\vspace{0.01in}
\noindent\textbf{Evaluation on different face manipulations.}
As mentioned earlier, the performance of recent DF detection methods is near-perfect (close to 99\% AUC) when evaluated on FF++ test set videos (after training on FF++ train set).
This also holds true to some extent even for unseen facial manipulation algorithms (such as for FSh, DFor), when the videos being evaluated are from the same dataset as training (FF++).
This can point to strong dependence on the training domain.
Therefore, such a high performance gain can be considered reliable when it also leads to improved cross-dataset and distortion/adversarial robustness.
Tab.~\ref{tab:distortions} shows the results for this cross-manipulation generalization analysis. 
PhaseForensics is amongst the top 3 for this set of evaluations, with a cross-manipulation generalization of at least $97.4\%$ AUC, only a $1.4\%$ drop compared to FTCN.
However, considering that the cross-dataset tests are crucial for understanding the real-world generalizability of DF detection, the performance gain of PhaseForensics over FTCN across three cross-dataset tests (Tab.~\ref{tab:main_results}a) outweighs this small drop in performance for within-domain results.

\vspace{0.05in}
\noindent\textbf{Robustness to spatial distortions.}
Color filters and compression are commonly applied to internet media. 
Here we assess robustness to such spatial distortions for PhaseForensics, and compare it to that of two existing temporal DF detectors, CNN-GRU and LipForensics.
We only consider distortions which do not dramatically alter the video appearance.
Here, amongst the distortions enlisted in the DFor dataset, we assess the color-based and compression-based distortions, at severity levels of $1$ and $2$. 
When a distortion to the video is perceptually very evident, it already points to an evidence of tampering of the video making the analysis of whether the video contains a deepfake possibly unnecessary. We defer the analysis on such obvious distortions and also of other distortion types (which are not as common) to the supplementary material.
Many previous works show the distortion robustness results on FF++ test set (the same DF generators are used for training)~\cite{Hal21}.
Here we analyze the distortion robustness on the test dataset of CDFv2~(Tab.~\ref{tab:distortions}) -- which is represents a more challenging real world use case, give the high quality DF generator and no overlap with training set.
PhaseForensics shows consistently higher robustness to color-change distortions (saturation and contrast change) and per-frame pixelation.
The performance for PhaseForensics is lower for compression artifacts, since these can affect the frequency composition of a video frame (discussed further in Sec.~\ref{subsec:ablation}).
However, PhaseForensics remains within a few percentage point of LipForensics for this setting.
Note that for this experiment, along with \%AUC, we also report mean absolute percentage error (MAPE) in AUC compared to the performance of each method on clean CDFv2.
A lower MAPE indicates that the method maintains its performance despite the applied distortion.
The performance of PhaseForensics is the least affected by distortions, except for compression.

\begin{figure}[t!]
	\centering
    \vspace{0.05in}
	\includegraphics[width=0.95\columnwidth, trim=0cm 0cm 0cm 0cm, clip=true]{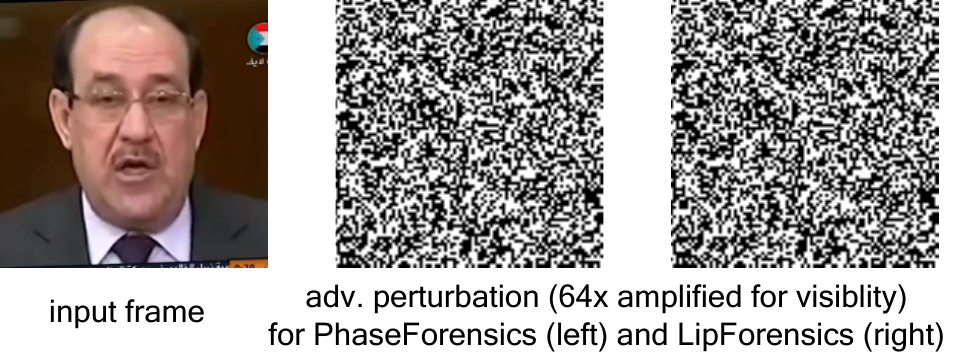}
	\vspace{-0.05in}
	\caption{\textbf{Examples of adversarial perturbations.} Here we show one example of an adversarial perturbation map for the black-box attack~\cite{Hus21} on LipForensics~\cite{Hal21} and PhaseForensics ($64\times$ amplified for visualization). These imperceptible noise signals can be added to the video, to fool the DF detection method.}
	\label{fig:adv_attack}
	\vspace{-0.2in}
\end{figure}

\begin{table*}[t!]
    \centering
    \vspace{0.1in}
    \centering
\begin{center}
	\begin{tabular}[b]{ m{1.2cm} | m{1.2cm} | m{1.8cm} | m{1cm} | m{1.2cm} |m{1.2cm} |m{1.2cm}}

\Xhline{2\arrayrulewidth}
%\textbf{METHOD} &  FF++~\cite{Ros19} & FSh~\cite{LiL20A}  & DFor~\cite{Jia20} & DFDC~\cite{Dol20} & VFHQ~\cite{Fox21} &
\tabledarkgray
\multicolumn{4}{c|}{CSP configuration}&\multicolumn{3}{c}{test dataset}\\
\Xhline{2\arrayrulewidth}
orient. & scales & bandwidth & input & CDFv2  & FF++ & DFor\\
\Xhline{2\arrayrulewidth}
\hline
2& 4 & octave& amp. & 66.1 & \textbf{99.5} & 97.3\\
\tablelightgray
2& 4 & half-octave & phase & 62.5 & \textbf{99.5} & 94.8 \\
% 2& 8 & half-octave & phase & &  & \\
\tablelightgray
2& 6 & octave & phase & 80.5 & 99.2 & 97.1 \\
4& 4 & octave & phase & 88.9 & \textbf{99.5} & 95.6\\
\tablelightgray
2& 4 & octave & phase & \textbf{91.2} & \textbf{99.5} & \textbf{97.4}\\
\Xhline{2\arrayrulewidth}
\end{tabular}
\end{center}

%97.3
%94.8
%
%97.1
%95.6
%97.4
    \vspace{-0.1in}
    \caption{We motivate the design choice of our phase-based input in PhaseForensics by training DF detectors with alternate input feature choices. We vary the CSP filter bandwidths, number of scales and orientations, and also demonstrate the result of training with purely amplitude-based input (in a similar spirit to existing works that use image-pyramid inputs~\cite{Mas20}). As is evident from \%AUC reported here for CDFv2, FF++ test set, and DFor, phase-based inputs from our choice of 4-scale, 2-orientation, octave-bandwidth pyramid prove to be the most optimal amongst the alternatives.}
    \vspace{-0.15in}
    \label{tab:csp_design}
\end{table*}

\noindent\textbf{Adversarial robustness.}
Recently, existing works have observed the vulnerability of DF detectors to black-box adversarial attacks~\cite{Hus21}.
Black-box attacks represent a plausible case where the detection model may be unknown to the adversary, but the adversary has the ability to test the model on a limited number of inputs.
To assess the adversarial robustness for CNN-GRU, LipForensics and PhaseForensics, we compute the adversarial perturbations (Fig.~\ref{fig:adv_attack}) for CDFv2 (and, in supplementary, on FSh) using a recent approach on black-box attacks for DF detectors (using the default settings provided by the authors)~\cite{Hus21}.
The \% AUC on adversarially-perturbed CDFv2 is shown in the last row of Tab.~\ref{tab:distortions}.
This black-box method uses Natural Evolutionary Strategies to estimate the gradient in each step, optimizing the adversarial image using an Expectation over Transforms and includes slight transformations of the input (blurring, translation) to the optimization loop for better robustness.
LipForensics relies on pixel intensity-based input without any constraints on the frequency components used for DF prediction.
Consequently, this leads to a higher susceptibility to adversarial attacks -- since typically such attacks reside in higher frequency bands~\cite{Wang20B}.
In contrast, PhaseForensics shows better robustness since the band-pass phase is used (excluding the high-frequency components).
Details of the attack parameters are available in the supplementary.
%!TEX root = ../main.tex

\subsection{Discussion and Ablation Studies}
\label{subsec:ablation}

\noindent\textbf{Design choices for CSP.} As stated in Sec.~\ref{sec:method}, the CSP for lip region is computed with 4 scales and 2 orientations, and the phase from complex band-pass coefficients is used (the real-valued low and high-pass residues do not contain phase) for training.
In Tab.~\ref{tab:csp_design}, we consider alternate input features to verify this choice.
We train DF detectors with input features computed from different configurations of CSP. 
Reducing the filter bandwidth (e.g., in Tab.~\ref{tab:csp_design}, from octave to half-octave) lowers the performance of DF detector since it increases the spatial support of the CSP filters~\cite{Wad13}, thereby reducing the fidelity to local motion.
Constructing a CSP with more than $4$ scales, for the lip sub-images of size $88\times88$, also does not yield an improvement -- and a similar observation holds for more than $2$ orientations.
Moreover, the third row shows a case where the phase information from bandpass components is discarded altogether: only the amplitude of the complex bandpass coefficients and the real-valued high and low-pass residues are used (similar, in spirit, to DF detection methods that use image pyramids with all frequency components~\cite{Mas20}).
In Tab.~\ref{tab:csp_design}, we can see that not relying on phase especially impacts cross-dataset generalization.
Across all experiments, the performance gain with the chosen octave-bandwidth, 4-scale, 2-orientation CSP is most evident in cross-dataset tests.

\begin{table}[t!]
    \centering
    \hspace{-0.25in}	
	\begin{minipage}{0.5\columnwidth}
        \hspace{-0.1in}	
        \centering
		\resizebox{0.8\columnwidth}{!}{\vspace{0.3in}\centering

	\begin{tabular}[b]{ m{1cm} | m{1cm} |m{1cm}}
	
	\Xhline{2\arrayrulewidth}
	%\textbf{METHOD} &  FF++~\cite{Ros19} & FSh~\cite{LiL20A}  & DFor~\cite{Jia20} & DFDC~\cite{Dol20} & VFHQ~\cite{Fox21} &
	\tabledarkgray
	region & CDFv2 &DFDC\\
	\hline
	eye  & 73.3 & 65.5\\
	\tablelightgray
	lip & 91.2& 78.2  \\
	\Xhline{2\arrayrulewidth}
\end{tabular}
}
		\resizebox{0.65\columnwidth}{!}{\vspace{0.in}(a) effect of face regions}
	\end{minipage}%
	\hspace{-0.2in}
	\begin{minipage}{0.5\columnwidth}
		\centering
		\resizebox{1.1\columnwidth}{!}{\centering

	\begin{tabular}[b]{ m{3cm} | m{1cm} |m{1cm}}

\Xhline{2\arrayrulewidth}
%\textbf{METHOD} &  FF++~\cite{Ros19} & FSh~\cite{LiL20A}  & DFor~\cite{Jia20} & DFDC~\cite{Dol20} & VFHQ~\cite{Fox21} &
\tabledarkgray
temporal processing & CDFv2 &DFDC\\
\hline
no processing  & 62.8 & 64.6\\
\tablelightgray
3DCNN  & 85.2& 77.3\\
separable 3DCNN & 91.2& 78.2  \\
\Xhline{2\arrayrulewidth}
\end{tabular}
}
		\vspace{0.1in}
		\resizebox{0.9\columnwidth}{!}{(b) choice of temporal processing}
	\end{minipage}
	\vspace{-0.1in}
    \caption{(a) Here we show that applying PhaseForensics to lip sub-region yields a better performance compared to eyes, since motion around eyes is less prominent compared to that around lips.
    (b) To verify our choice of temporal processing of phase using separable 3DCNN, we train three models with different types of temporal processing. First, we apply no processing and feed the phase directly to the ResNet feature extractor. Second, we apply a 3D conv layer to phase. Lastly, we apply a separable 3D convolution layer: which proves to be the most optimal choice.}
    \vspace{-0.18in}
    \label{tab:eye_vs_lip}
\end{table}

\vspace{0.05in}
\noindent\textbf{PhaseForensics applied to eyes.}
While lips provide a strong motion cue, in Tab.~\ref{tab:eye_vs_lip}a, we evaluate PhaseForensics applied to lips, after pretraining on gaze prediction task (Sec.~\ref{sec:method}).
As expected, learning a DF detector from eye regions using phase changes is difficult compared to lips, due to the nature of eye motion: more sporadic, and low amplitude compared to lips. See supplementary material for experiment on training with full face as input.

\vspace{0.05in}
\noindent\textbf{Temporal preprocessing.}
To motivate the importance of separable spatio-temporal phase prepossessing (Sec.~\ref{sec:method}), we train three models on the lip region: one without any spatio-temporal filtering (phase information is directly passed to ResNet-18 feature extractor), one with a standard 3DCNN layer (similar to ~\cite{Hal21}), and one with our proposed separable 3D CNN layer. Looking at the performance on CDFv2, DFDC datasets in Tab.~\ref{tab:eye_vs_lip}b, we conclude that our choice of separable spatio-temporal phase processing is most suited for PhaseForensics.

\vspace{0.05in}
\noindent\textbf{Benefit of phase in comparison to pixel-intensity inputs.}
Note that, the second row Tab.~\ref{tab:eye_vs_lip}b (3DCNN) is a direct modification of the LipForensics pipeline: instead of training with pixel intensities as input (as done in LipForensics), we feed $\Phi(t,x,y,c)$ from Eq.~\ref{eq:phase_postproc} to their model, appropriately adjusting the number of input channels.
This allows for a clear assessment of the advantage of using phase instead of pixel values: compared to LipForensics, using phase inputs in the LipForensics pipeline improves the \%AUC on CDFv2 from 82.4 to 85.2. As discussed above, further improvement is obtained by adopting Eq.~\ref{eq:phase_postproc}.

\vspace{0.05in}
\noindent\textbf{Limitations.}
Similar to LipForensics~\cite{Hal21}, PhaseForensics is dependent on the presence of motion in the lips and requires a domain-specific pretraining step.
Moreover, CSP decompositions currently tend to form overcomplete/redundant representations for images, which may create additional computational cost. 
More efficient alternatives such as Reisz pyramids~\cite{Wad14} could be used in the future.
Lastly, distortion and adversarial robustness with PhaseForensics holds so long as the perturbations do not disrupt the band-passed frequency components.
Therefore, the adversarial robustness of PhaseForensics depends on careful filter design, to ensure that the final model is not susceptible to high-frequency attacks.

%!TEX root = ../main.tex
\section{Conclusion}
In this work, we presented PhaseForensics, our method for generalizable, robust, deepfake detection.
This method outperforms existing works with state-of-the art cross-dataset generalization, and is robust under a variety of spatial and adversarial distortions.
By being effective against both in and out of training domain samples, we take a step towards the real-world deployment of DF detectors.

\vspace{-0.05in}
\noindent \paragraph{Acknowledgements.} We would like to thank Seanwook Park for help with setting up the EVE dataset, Alexandros Haliassos for discussion and clarifications on LipForensics, and Abhishek Badki and Orazio Gallo for the initial discussions. Ekta Prashnani was supported in part by a Dissertation Fellowship provided by the Department of Electrical and Computer Engineering at UC Santa Barbara. This work was supported in part through NSF SI2-SS1 award no. 1664172 and through the computational facilities purchased using NSF grant OAC-1925717 and administered by the Center for Scientific Computing (CSC) at UC Santa Barbara. The CSC is supported by the California NanoSystems Institute and the Materials Research Science and Engineering Center (MRSEC; NSF DMR 1720256).

{\small
\bibliographystyle{ieee_fullname}
\bibliography{main}
}

\newpage
\newpage
\clearpage
\twocolumn[
  \begin{@twocolumnfalse}
{
   \newpage
   \null
   \begin{center}
      {\Large \bf {G}eneralizable {D}eepfake {D}etection with {P}hase-{B}ased {M}otion {A}nalysis \\ ({S}upplementary)}
      {
      \large
      \lineskip .5em
      \begin{tabular}[t]{c}
          
      \end{tabular}
      \par
      }
      \vskip .5em
      \vspace*{0pt}
   \end{center}
}
  \end{@twocolumnfalse}
]

%!TEX root = ../main.tex
\section{Training details (Sec.~3 continued)}
\label{sec:sup_train}

\noindent\textbf{Preprocessing and data augmentation steps.}
We use RetinaFace for detecting facial sub-regions~\cite{Den20}, and FAN~\cite{Bul17} for landmark detection.
Facial region crops ($1.3\times$ larger than detected face region~\cite{Ros19}) are then aligned by affine warping each frame so that $5$ landmarks around eyes and nose match an average face, along with temporal smoothing over $12$ frames for the landmarks to avoid jitter.
This preprocessing is same as a previous work: LipForensics~\cite{Hal21}.

The specified facial sub-region (eyes or lips) is then extracted from aligned facial crops. In case of eyes, an initial crop of size $64\times128$ is performed centered around both eyes. 
From this, a random crop of size $56\times122$ is selected for training.
For lips, an initial crop of $96\times96$ is used, with a random crop of size $88\times88$ used for training.
In case of lips, random horizontal flipping is also performed for the training set.

\vspace{0.1in}
\noindent\textbf{More architecture details.} 
The input clip-size is $25$ frames~\cite{Hal21}, which is passed to the complex-steerable pyramid (CSP) computation step~\cite{Por00}\footnote{\href{https://github.com/tomrunia/PyTorchSteerablePyramid}{https://github.com/tomrunia/PyTorchSteerablePyramid}}.
As mentioned in the main paper, the phase band-pass components of the CSP are appended along the channel dimension, making the input being passed to $s_{x,y}$ (Eq.~2 of main paper) of size $32 \times C \times 25 \times H\times W$ (in terms of the PyTorch $NCDHW$ convention for $3D$ inputs), where our batch size is set to $32$, $C$ depends on the number of scales and orientations of the bandpass coefficients (specified in the main paper for eyes and lips) and $(H,W)$ are the spatial dimensions of the facial subregion (mentioned above).
The function $s_{x,y}$ is implemented as a 3DCNN operation with a filter of size $1\times7\times7$ (first dimension filters along the time axis) with a stride of $2$, and padding $3$ followed by batch normalization and ReLU non-linearity~\cite{Xie18} --with $64$ output channels.
The function $f_{t}$ is implemented as a 3DCNN operation with a filter of size $3\times1\times1$ with a stride of $1$ and padding $1$, followed by batch normalization and ReLU non-linearity~\cite{Xie18} --with $64$ output channels.
The remainder of the model consists of a standard ResNet-18 feature extractor, followed by an MS-TCN adopted from the previous work on sequence modeling for lip reading and deepfake (DF) detection~\cite{Mar20,Hal21} (Fig.~2 of main paper).

\vspace{0.1in}
\noindent\textbf{Pretraining details.} 
As mentioned in the main paper, instead of starting from random weights or with ImageNet-pretrained weights (as is often done), we pretrain the DF detector on tasks specific to the facial regions.

When training with eyes sub-region, the DF detector is finetuned on pretrained weights obtained from gaze detection task with the EVE dataset~\cite{Par20}, where we train for $50,000$ iterations to obtain a small validation-set performance of $3.6$ angular loss.

For lip region, the pretraining with LRW dataset~\cite{Chu16} is performed for $10$ epochs, yielding a validation-set accuracy of $77\%$ (early stopping).
The output layer from pretrained networks is switched out for a $1$-class linear classifier layer before finetuning for DF detection.
The finetuning step is performed using FF++ training set, with the performance (in terms of loss) on FF++ validation set used to identify the best model, which is then evaluated on different datasets for generating all the results.
The choice for number of frames for training, validation and test set is consistent with LipForensics~\cite{Hal21}.

%!TEX root = ../main.tex
\section{Results (Sec.~4 continued)}
\label{sec:sup_results}

\noindent\textbf{Robustness to spatial distortions.} 
We continue our analysis of robustness to spatial distortions for the three baselines (CNN-GRU~\cite{Sab19}, LipForensics~\cite{Hal21} and our proposed method, PhaseForensics) mentioned in the main paper (Sec.~4), on the distortions enlisted in DFor datasets, applied to CDFv2 (high-quality faceswap deepfakes~\cite{Li20}) and also to an additional dataset: VFHQ (high-quality face re-enactment deepfakes~\cite{Fox21}).
\begin{figure*}
	\centering
	\includegraphics[width=1.0\linewidth, trim=0cm 0cm 0cm 0cm, clip=true]{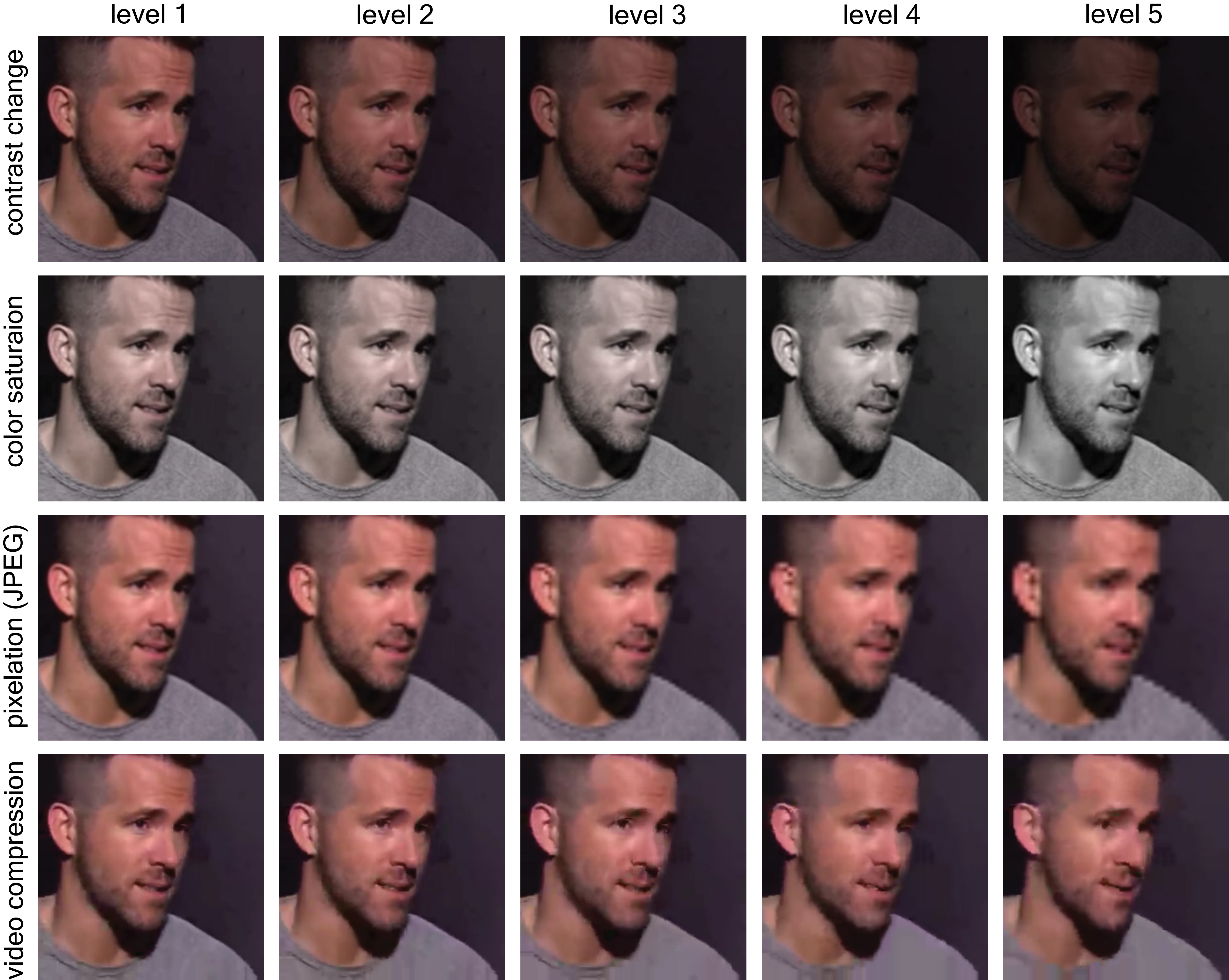}
	\caption{\textbf{Visualization of the spatial distortions that are routinely applied to internet media.} Here we visualize a frame from the CDFv2 dataset with various color-based and compression-based distortions applied (based on DFor distortion code). These distortions present a threat to the DF detectors, since they do not leave clear signs of tampering -- given their ubiquity -- and adversely affect the performance of DF detectors (Tab.~\ref{tab:common_dist_cdfv2},~\ref{tab:common_dist_vfhq}).}
	\label{fig:common_dist}
\end{figure*}

\begin{table*}[!ht]
	\centering
	\centering
%\begin{tabular}{ m{2cm} | m{0.5cm} | m{0.5cm} | m{0.5cm} | m{0.5cm} | m{0.5cm} | m{0.5cm} | m{0.5cm} | m{0.5cm} | m{0.5cm} | m{0.5cm} | m{0.5cm} | m{0.5cm} | m{0.5cm}}
\scalebox{0.9}{
\begin{tabular}{ m{2cm} | m{0.8cm} | m{1.8cm} | m{1.8cm} | m{1.8cm} | m{1.8cm} | m{1.4cm} | m{1.8cm}}

\Xhline{2\arrayrulewidth}
\tabledarkgray
\multicolumn{2}{c|}{}&\multicolumn{6}{c}{\textbf{CDFv2}}\\
\hline
\multicolumn{2}{c|}{\textbf{Distortion}}&\multicolumn{2}{c|}{\textbf{CNN-GRU}}&\multicolumn{2}{c|}{\textbf{LipForensics}}&\multicolumn{2}{c}{\textbf{PhaseForensics}}\\
\Xhline{2\arrayrulewidth}
type & level & $\uparrow$\%auc & $\downarrow$ \%mape & $\uparrow$\%auc & $\downarrow$ \%mape & $\uparrow$\%auc & $\downarrow$ \%mape \\
\Xhline{2\arrayrulewidth}
\tablelightgray
&1&73.6&5.4&74.8&9.2&\textbf{89.5}&\textbf{1.9}\\
\tablelightgray
contrast&2&73.5&5.3&74.7&9.3&\textbf{90}&\textbf{1.3}\\
\tablelightgray
change&3&73.4&5.2&74.6&9.5&\textbf{89.4}&\textbf{2}\\
\tablelightgray
&4&73&4.6&74.2&10&\textbf{90.3}&\textbf{1}\\
\tablelightgray
&5&72.5&3.9&73.4&10.9&\textbf{89.1}&\textbf{2.3}\\

&1&70.5&1&72.8&11.7&\textbf{90.9}&\textbf{0.3}\\
color&2&69.4&0.6&72.2&12.4&\textbf{90.9}&\textbf{0.3}\\
saturation&3&67.8&2.9&71.7&13&\textbf{90.8}&\textbf{0.4}\\
&4&66&5.4&71.2&13.6&\textbf{90.6}&\textbf{0.7}\\
&5&63.9&8.5&70.1&14.9&\textbf{91.1}&\textbf{0.1}\\

\tablelightgray
&1&64.8&7.2&76.3&7.4&\textbf{91.3}&\textbf{0.1}\\
\tablelightgray
&2&62.7&10.2&74.6&9.5&\textbf{84.3}&\textbf{7.6}\\
\tablelightgray
pixelation&3&60.4&\textbf{13.5}&70.8&14.1&\textbf{78.7}&13.7\\
\tablelightgray
&4&60.4&\textbf{13.5}&67.9&17.6&\textbf{74.7}&18.1\\
\tablelightgray
&5&59.3&\textbf{15}&64.2&22.1&\textbf{67.9}&25.5\\

&1&70.5&\textbf{1}&\textbf{74.3}&9.8&72.6&20.4\\
&2&69.4&\textbf{0.6}&\textbf{72.2}&12.4&70.3&22.9\\
compression&3&66.4&\textbf{4.9}&\textbf{69}&16.3&64.6&29.2\\
&4&\textbf{64.1}&\textbf{8.2}&63.8&22.6&56.2&38.4\\
&5&\textbf{61.9}&\textbf{11.3}&61.1&25.8&50.1&45.1\\

\Xhline{2\arrayrulewidth}

\end{tabular}
}
	\caption{\textbf{Analyzing DF detection robustness to spatial distortions that are routinely observed on internet media (evaluated on CDFv2~\cite{Li20}):} Here we compare CNN-GRU~\cite{Sab19}, LipForensics~\cite{Hal21} and our proposed method, PhaseForensics, on commonly-occurring spatial distortions found on the internet media (color, compression)
	The distortion code used here is obtained from DFor dataset~\cite{Jia20} and applied to CDFv2 dataset.
	Overall, PhaseForensics outperforms other baselines on all color-based distortions and comparable or better than existing approaches on compression-based distortions. In addition to \%AUC, we also report the mean absolute percentage error (MAPE) in \%AUC compared to the performance of each method on undistorted CDFv2 (reported in Tab.~1a of the main paper).}
	\label{tab:common_dist_cdfv2}
\end{table*}
\begin{table*}[!ht]
	\centering
	\centering
%\begin{tabular}{ m{2cm} | m{0.5cm} | m{0.5cm} | m{0.5cm} | m{0.5cm} | m{0.5cm} | m{0.5cm} | m{0.5cm} | m{0.5cm} | m{0.5cm} | m{0.5cm} | m{0.5cm} | m{0.5cm} | m{0.5cm}}
\scalebox{0.9}{
\begin{tabular}{ m{2cm} | m{0.8cm} | m{1.8cm} | m{1.8cm} | m{1.8cm} | m{1.8cm} | m{1.8cm} | m{1.8cm}}

\Xhline{2\arrayrulewidth}
\tabledarkgray
\multicolumn{2}{c|}{} & \multicolumn{6}{c}{\textbf{VFHQ}}\\
\hline
\multicolumn{2}{c|}{\textbf{Distortion}} & \multicolumn{2}{c|}{\textbf{CNN-GRU}} & \multicolumn{2}{c|}{\textbf{LipForensics}} & \multicolumn{2}{c}{\textbf{PhaseForensics}}\\
\Xhline{2\arrayrulewidth}
type  &  level  &  $\uparrow$\%auc  &  $\downarrow$ \%mape  &  $\uparrow$\%auc  &  $\downarrow$ \%mape  &  $\uparrow$\%auc  &  $\downarrow$\%mape \\
\Xhline{2\arrayrulewidth}

\tablelightgray
 & 1 & 60 & 9.1 & 88.2 & 2.2 & \textbf{94.6} & \textbf{0.4}\\
\tablelightgray
contrast & 2 & 60.3 & 8.6 & 88 & 2.4 & \textbf{94.5} & \textbf{0.3}\\
\tablelightgray
change & 3 & 58.9 & 10.8 & 87.7 & 2.8 & \textbf{94.7} & \textbf{0.5}\\
\tablelightgray
 & 4 & 59.5 & 9.8 & 87.4 & 3.1 & \textbf{94.7} & \textbf{0.5}\\
\tablelightgray
 & 5 & 58.1 & 12 & 86.6 & 4 & \textbf{95.3} & \textbf{1.2}\\

 & 1 & 46.2 & 30 & 90.3 & 0.1 & \textbf{94.5} &\textbf{ 0.3}\\
color & 2 & 42 & 36.4 & 90.4 & 0.2 & \textbf{94.3} &\textbf{ 0.1}\\
saturation & 3 & 37.1 & 43.8 & 90.5 & 0.3 & \textbf{94.4} & \textbf{0.2}\\
 & 4 & 34.4 & 47.9 & 90.5 & 0.3 & \textbf{94.4} & \textbf{0.2}\\
 & 5 & 33.7 & 48.9 & 90.5 & \textbf{0.3}\textbf{} & \textbf{94.6} & 0.4\\

\tablelightgray
 & 1 & 60.9 & 7.7 & 96.5 & 7 & \textbf{92.2} & \textbf{2.1}\\
\tablelightgray
 & 2 & 59.5 & 9.8 & \textbf{96.4} & \textbf{6.9} & 72.9 & 22.6\\
\tablelightgray
pixelation & 3 & 52.4 & 20.6 & \textbf{94.2} & \textbf{4.4} & 77.1 & 18.2\\
\tablelightgray
 & 4 & 49.8 & 24.5 & \textbf{87.6} & \textbf{2.9} & 66.3 & 29.6\\
\tablelightgray
 & 5 & 38.3 & 42 & \textbf{83.8} & \textbf{7.1} & 61.4 & 34.8\\

 & 1 & 56.6 & 14.2 & \textbf{81.5} & \textbf{9.6} & 81.3 & 13.7\\
 & 2 & 54.1 & 18 & 74.9 & 17 & \textbf{78.9} & \textbf{16.2}\\
compression & 3 & 51.5 & 22 & 71.4 & \textbf{20.8} & \textbf{73.8} & 21.7\\
 & 4 & 52 & 21.2 & \textbf{66.9} & \textbf{25.8} & 66.7 & 29.2\\
 & 5 & 52.4 & 20.6 & 62.2 & \textbf{31} & \textbf{62.3} & 33.9\\

\Xhline{2\arrayrulewidth}

\end{tabular}
}
	\caption{\textbf{Analyzing DF detection robustness to spatial distortions that are routinely observed on internet media (evaluated on VFHQ~\cite{Fox21}):} Here we repeat the analysis from Tab.~\ref{tab:common_dist_cdfv2}, on VFHQ.}
	\label{tab:common_dist_vfhq}
\end{table*}

\begin{table*}[!ht]
	\centering
	\centering

\scalebox{0.9}{
\begin{tabular}{ m{2cm} | m{0.8cm} | m{1.8cm} | m{1.8cm} | m{1.8cm} | m{1.8cm} | m{1.8cm} | m{1.8cm}}

\Xhline{2\arrayrulewidth}
\tabledarkgray
\multicolumn{2}{c|}{}&\multicolumn{6}{c}{\textbf{CDFv2}}\\
\hline
\multicolumn{2}{c|}{\textbf{Distortion}}&\multicolumn{2}{c|}{\textbf{CNN-GRU}}&\multicolumn{2}{c|}{\textbf{LipForensics}}&\multicolumn{2}{c}{\textbf{PhaseForensics}}\\
\Xhline{2\arrayrulewidth}
type & level & $\uparrow$\%auc & $\downarrow$ \%mape & $\uparrow$\%auc & $\downarrow$ \%mape & $\uparrow$\%auc & $\downarrow$ \%mape \\
\Xhline{2\arrayrulewidth}

\tablelightgray
&1&73.6&5.4&73.1&11.3&\textbf{91}&\textbf{0.2}\\
\tablelightgray
blockwise&2&73.4&\textbf{5.2}&67.9&17.6&\textbf{79.6}&12.7\\
\tablelightgray
distortion&3&\textbf{72.2}&\textbf{3.4}&66.4&19.4&65.9&27.7\\
\tablelightgray
&4&\textbf{72}&\textbf{3.2}&64.4&21.8&53.6&41.2\\
\tablelightgray
&5&\textbf{71.3}&\textbf{2.1}&62.2&24.5&43.7&52.1\\

&1&55.7&\textbf{20.2}&\textbf{62.1}&24.6&46.1&49.5\\
Gaussian&2&51.3&26.5&\textbf{65}&\textbf{21.1}&38.8&57.5\\
noise&3&50.5&27.7&\textbf{67.5}&\textbf{18.1}&35.9&60.6\\
&4&56.2&\textbf{19.5}&\textbf{65}&21.1&36.7&59.8\\
&5&54&\textbf{22.6}&\textbf{57.9}&29.7&35.9&60.6\\

\tablelightgray
&1&61.5&11.9&74&\textbf{10.2}&\textbf{80}&12.3\\
\tablelightgray
&2&58.1&16.8&\textbf{69.2}&\textbf{16}&54.9&39.8\\
\tablelightgray
Gaussian blur&3&54.3&\textbf{22.2}&\textbf{61.4}&25.5&51.5&43.5\\
\tablelightgray
&4&52.3&\textbf{25.1}&\textbf{56.2}&31.8&55.4&39.3\\
\tablelightgray
&5&51.2&\textbf{26.6}&52.8&35.9&\textbf{55.3}&39.4\\

\Xhline{2\arrayrulewidth}

\end{tabular}
}
	\caption{\textbf{Robustness to other spatial distortions that show clear signs of tampering (evaluation on CDFv2~\cite{Li20}).} Here we compare the performance of CNN-GRU, LipForensics, and PhaseForensics for $3$ additional distortions at all severity levels: random block-wise perturbations, gaussian noise and gaussian blur (visualized in Fig.~\ref{fig:less_common_dist}); using code from DFor dataset~\cite{Jia20}). Except for very low levels, these distortions leave clear evidence of tampering -- making them less likely to be deployed as attacks used to confuse DF detectors. Overall, the results are mixed for these kinds of less common distortions -- with no clear winner for any particular type of distortion. In addition to \%AUC, we also report the mean absolute percentage error (MAPE) in \%AUC compared to the performance of each method on undistorted CDFv2 (reported in Tab.~1a of the main paper).}
	\label{tab:uncommon_dist_cdfv2}
	\vspace{-0.1in}
\end{table*}
\begin{table*}[!ht]
	\centering
	\centering
%\begin{tabular}{ m{2cm} | m{0.5cm} | m{0.5cm} | m{0.5cm} | m{0.5cm} | m{0.5cm} | m{0.5cm} | m{0.5cm} | m{0.5cm} | m{0.5cm} | m{0.5cm} | m{0.5cm} | m{0.5cm} | m{0.5cm}}
\scalebox{0.9}{
\begin{tabular}{ m{2cm} | m{0.8cm} | m{1.8cm} | m{1.8cm} | m{1.8cm} | m{1.8cm} | m{1.8cm} | m{1.8cm}}

\Xhline{2\arrayrulewidth}
\tabledarkgray
\multicolumn{2}{c|}{}& \multicolumn{6}{c}{\textbf{VFHQ}}\\
\hline
\multicolumn{2}{c|}{\textbf{Distortion}}& \multicolumn{2}{c|}{\textbf{CNN-GRU}}& \multicolumn{2}{c|}{\textbf{LipForensics}}& \multicolumn{2}{c}{\textbf{PhaseForensics}}\\
\Xhline{2\arrayrulewidth}
type &  level &  $\uparrow$\%auc &  $\downarrow$ \%mape &  $\uparrow$\%auc &  $\downarrow$ \%mape &  $\uparrow$\%auc &  $\downarrow$ \%mape \\
\Xhline{2\arrayrulewidth}

\tablelightgray
& 1& 58.7& 11.1& \textbf{90.1}&\textbf{ 0.1}& 84.7& 10.1\\
\tablelightgray
blockwise& 2& 56& 15.2& \textbf{89}& \textbf{1.3}& 77.4& 17.8\\
\tablelightgray
distortion& 3& 55.2& 16.4& \textbf{85.2}& \textbf{5.5}& 65.2& 30.8\\
\tablelightgray
& 4& 52.6& 20.3& \textbf{78.1}& \textbf{13.4}& 59& 37.4\\
\tablelightgray
& 5& 49.5& 25& \textbf{73.6}& \textbf{18.4}& 52.1& 44.7\\

& 1& 38.8& 41.2& 46.3& 48.7& \textbf{80.1}& \textbf{15}\\
Gaussian& 2& 38.3& 42& 61.1& 32.3& \textbf{70.3}& \textbf{25.4}\\
noise& 3& 45.1& 31.7& \textbf{71.4}& \textbf{20.8}& 49.2& 47.8\\
& 4& 25.2& 61.8& \textbf{69.4}& \textbf{23.1}& 44.6& 52.7\\
& 5& 22.2& 66.4& \textbf{58.7}& \textbf{34.9}& 45.4& 51.8\\

\tablelightgray
& 1& 51.6& 21.8& \textbf{96}& 6.4& 93.5& \textbf{0.7}\\
\tablelightgray
& 2& 46.6& 29.4& \textbf{95.4}& \textbf{5.8}& 87.5& 7.1\\
\tablelightgray
Gaussian blur& 3& 32.6& 50.6& \textbf{88.4}& \textbf{2}& 49.3& 47.7\\
\tablelightgray
& 4& 31.9& 51.7& \textbf{85.2}& \textbf{5.5}& 55.8& 40.8\\
\tablelightgray
& 5& 30.5& 53.8& \textbf{80.3}& \textbf{11}& 62.9& 33.2\\

\Xhline{2\arrayrulewidth}

\end{tabular}
}
	\caption{\textbf{Robustness to other spatial distortions that show clear signs of tampering  (evaluation on VFHQ~\cite{Fox21}).} Here we compare the performance of CNN-GRU, LipForensics, and PhaseForensics for $3$ additional distortions at all severity levels: random block-wise perturbations, gaussian noise and gaussian blur (visualized in Fig.~\ref{fig:less_common_dist}); using code from DFor dataset~\cite{Jia20}), applied to VFHQ. In addition to \%AUC, we also report the mean absolute percentage error (MAPE) in \%AUC compared to the performance of each method on undistorted VFHQ (reported in Tab.~1a of the main paper).}
	\label{tab:uncommon_dist_vfhq}
	\vspace{-0.1in}
\end{table*}

\begin{figure*}
	\centering
	\includegraphics[width=1.0\linewidth, trim=0cm 0cm 0cm 0cm, clip=true]{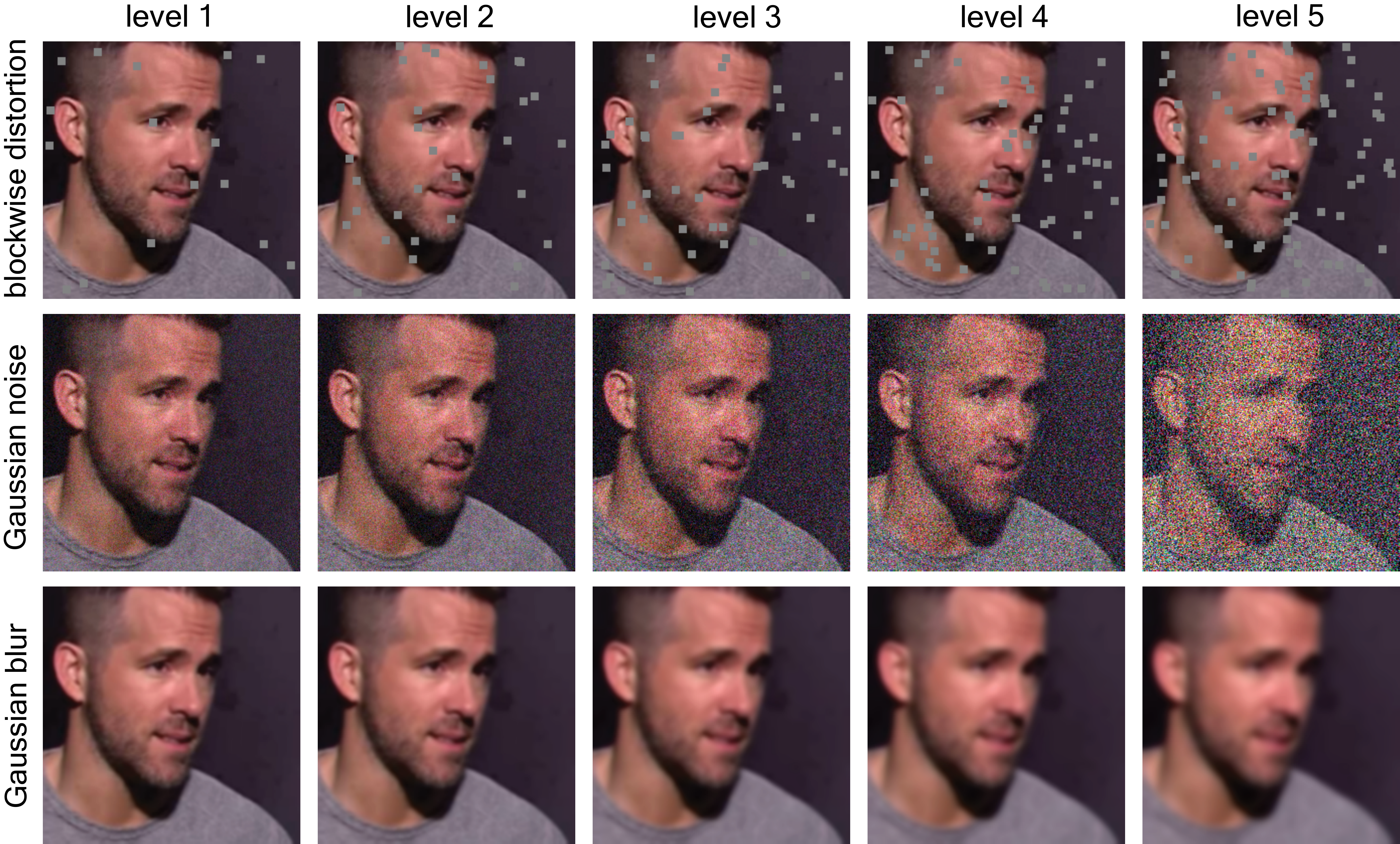}
	\caption{Visualization of the spatial distortions that are less common and leave clear evidence of tampering.}
	\label{fig:less_common_dist}
\vspace{-0.1in}
\end{figure*}

In Tab.~\ref{tab:common_dist_cdfv2},\ref{tab:common_dist_vfhq}, we compare the baselines on commonly-occurring spatial distortions found on the internet media (color, compression) that do not raise a suspicion about the authenticity of the media, given their ubiquity, on CDFv2 and VFHQ datasets respectively.  We want to assess the cross-dataset generalizability of the the DF detection methods in the presence of such distortions.
For color-based distortions, PhaseForensics is particularly effective.
For per-frame compression-based distortions (JPEG compression -- referred here as pixelation consistent with~\cite{Hal21}), PhaseForensics is more effective for CDFv2 (face-swap deepfakes), as compared to VFHQ (face re-enactment deepfakes).
For video compression, LipForensics and PhaseForensics are comparable for VFHQ, while for CDFv2 -- PhaseForensics is slightly worse than LipForensics for low compression levels.
Fig.~\ref{fig:common_dist} visualizes the distortions considered here for a video frame from CDFv2 dataset.
The code from the DFor dataset~\cite{Jia20} is used for all of the distortion robustness analysis.
Higher levels of distortions indicate clear signs of tampering -- reducing the likelihood of being deployed adversarially to fool DF detectors.
Overall, PhaseForensics performance is better on all color-based distortions and comparable or better than LipForensics on compression-based distortions.
The same trend holds with mean absolute percentage error.

\begin{table*}[!h]
	\centering
	\centering
	\begin{tabular}[b]{ m{3cm} | m{2cm} |m{2cm} | m{2cm} |m{2cm}}
	
	\Xhline{2\arrayrulewidth}
	%\textbf{METHOD} &  FF++~\cite{Ros19} & FSh~\cite{LiL20A}  & DFor~\cite{Jia20} & DFDC~\cite{Dol20} & VFHQ~\cite{Fox21} &
	\tabledarkgray
	model & \multicolumn{2}{c|}{\textbf{CDFv2}} & \multicolumn{2}{c}{\textbf{FSh}}\\
	\tabledarkgray
	& \%auc $\uparrow$ & \%mape $\downarrow$ & \%auc $\uparrow$ & \%mape $\downarrow$ \\
	\hline
	CNN-GRU  & 43.4 & 37.8 & 49.7 & 38.5\\
	\tablelightgray
	LipForensics  & 49.5 & 39.9 & 62.7 & 35.4\\
	PhaseForensics  & \textbf{71.1} & 22.0 & \textbf{69.7} & \textbf{28.4}\\
\Xhline{2\arrayrulewidth}	
\end{tabular}

	\caption{\textbf{Robustness to adversarial perturbations.} PhaseForensics is particularly advantageous in this regard, given its dependence on bandpass frequency components.}
	\label{tab:more_adv_robustness}
	\vspace{-0.2in}
\end{table*}

In Tab.~\ref{tab:uncommon_dist_cdfv2},\ref{tab:uncommon_dist_vfhq}, we compare the robustness of DF detectors to other spatial distortions that show clear signs of tampering, on CDFv2 and VFHQ respectively. We compare the performance of CNN-GRU, LipForensics, and PhaseForensics for the $3$ distortions at all severity levels: random block-wise perturbations, gaussian noise and gaussian blur (visualized in Fig.~\ref{fig:less_common_dist}; using code from DFor dataset~\cite{Jia20}). Except for very low levels, these distortions leave clear evidence of tampering -- making them less likely to be deployed as attacks used to confuse DF detectors. However, for completeness, we compare performance on these distortions as well. In case of Gaussian noise, LipForensics performs better in terms of AUC for CDFv2, while PhaseForensics performs better on VFHQ, at low distortion levels. Block-wise distortions perhaps leave the most evident signs of tampering: with random colored blocks appearing at different locations for each frame. For this distortion, at low levels, PhaseForensics performs best for CDFv2, while LipForensics performs best for VFHQ. Admittedly, PhaseForensics is most vulnerable to high amount of Gaussian blur -- since it distorts frequency distribution. This manifests as worse performance with PhaseForensics, as compared to LipForensics, in all cases except for a low distortion level for CDFv2. A future step to overcome this could involve learning a better frequency decomposition for the input frames. Overall, the results are mixed for these kinds of less common distortions.

\vspace{0.05in}
\noindent\textbf{Robustness to adversarial distortions.} 
We extend our analysis on the robustness of CNN-GRU and LipForensics (two of the baselines for temporal DF detection methods~\cite{Sab19,Hal21}) to black box attacks~\cite{Hus21} on FSh~\cite{Li20}, in addition to the CDFv2 analysis presented in Tab.~2 of the main paper (see Tab.~\ref{tab:more_adv_robustness}).
This black-box method utilizes Natural Evolutionary Strategies (NES) to estimate gradients between the input and output of the network without knowing the model weights.
This is done by applying small perturbations to the input image in the form of Gaussian noise, and observing the change in output result.
With an input image scaled to the range of [0,1], we use 40 samples per step, and take a step size of 1/255.
At most 25 total steps are taken, with a total perturbation of at most 16/255.
Consistent with the original work~\cite{Hus21}, once the adversarial perturbation reaches a 90\% confidence of fooling the detector, an Expectation over Transforms is introduced.
This step applies random Gaussian blur and translation to the video before each adversarial step.

Consistent with our results in the main paper, PhaseForensics continues to outperforms these methods as well.
As discussed in the main paper, PhaseForensics shows better adversarial robustness since the phase computations are performed in band-pass frequency components -- reducing the chance of being affected by adversarial attacks. 
In contrast, other methods rely on pixel intensity-based input without any constraints on the frequency components which enable the DF prediction.

\begin{table}[!h]
	\centering
	\centering
\begin{center}
	\begin{tabular}[b]{ m{3cm} | m{1cm} |m{1cm}}
	
	\Xhline{2\arrayrulewidth}
	%\textbf{METHOD} &  FF++~\cite{Ros19} & FSh~\cite{LiL20A}  & DFor~\cite{Jia20} & DFDC~\cite{Dol20} & VFHQ~\cite{Fox21} &
	\tabledarkgray
	pre-training & CDFv2 &DFDC\\
	\hline
	no pretraining  & 76.4 & 61.8\\
	\tablelightgray
	lipreading pretrained  & 91.2 & 78.2\\
	\Xhline{2\arrayrulewidth}
\end{tabular}
\end{center}
	\caption{\textbf{Effectiveness of pretraining on lip reading task.} Here we show the performance of PhaseForensics (\%AUC), with and without the lip reading pretraining, on CDFv2 and DFDC. As is clear, LipReading pretraining is crucial for effective DF detection.}
	\label{tab:lipread_pretrain}
	\vspace{-0.2in}
\end{table}

\vspace{0.1in}
\noindent\textbf{Performance of PhaseForensics on lips, with and without pretraining on lipreading task.}
As detailed in Sec.~3.2 of the main paper, similar to LipForensics~\cite{Hal21}, pretraining on lipreading (using LRW dataset~\cite{Chu16}, Sec.~3.2 of main paper) enables us to learn a distribution of natural lip movements. Finetuning such a pretrained network allows us to effectively train a DF classifier by detecting deviations from natural lip movements. As shown in Tab.~\ref{tab:lipread_pretrain}, the performance of the PhaseForensics DF detector drops without the pretraining step. This is expected, since such a case is akin to expecting the DF detector to predict deviations from natural lip movements, without having seen the original distribution of natural lip movements first.

\vspace{0.1in}
\noindent\textbf{Additional experiments with PhaseForensics.}
In Tab.~\ref{tab:more_exp}, we report the following additional experiments that justify the current design choices for PhaseForensics.
\begin{enumerate}
    \item \textbf{Using phase from DFT instead of CSP:} A CSP, given its finite support basis, isolates spatially-localized motion in its phase changes -- which is needed for local facial motion processing. In contrast, the phase changes with DFT capture the global motion. An experiment using DFT phase is shown below: using CSP phase (last row) provides better results.
    \item \textbf{Training on the full face:}  We thank the reviewer for pointing out the full-face training result, which we now show below: using face sub-regions (e.g. lips) compares favorably, since it allows the model to only look at relevant areas.
    \item \textbf{Choice of pre-training dataset:}  Since PhaseForensics relies on video-based input, we pretrain on Kinetics-400 action-recognition dataset as an alternative; this is also done in LipForensics~\cite{Hal21}). We observe that this leads to sub-optimal results (below): our original choice of domain-specific pre-training (e.g., lip-reading dataset for lip-based DF detection) allows the neural net to first learn from the realistic facial motion and then finetuning for DF detection allows to accurately detect DFs. Same as LipForensics~\cite{Hal21}, we find this anomaly-detection training approach to be more effective.
\end{enumerate}
\vspace{-0.2in}
\begin{table}[!h]
	\centering
	\resizebox{0.98\columnwidth}{!}{\begin{tabular}[b]{c|c|c|c}
&\multicolumn{3}{c}{\%AUC on test datasets of:}\\\cline{2-4}
different methods / training settings & FF++ test & FSh & CDFv2\\
\cline{1-4}
Lip region, DFT Phase, pretain on lip-reading & 99.1 & 95.7 & 66.6\\
Lip region, CSP Phase, Kinetics pretain on lip-reading &84.2 &60.9 &77.0 \\
Face region, CSP Phase, pretrain on lip-reading &98.3 &92.9 &81.6 \\
\textbf{PhaseForensics: lip, CSP Phase, lip-reading pretrain} &\textbf{99.5} &\textbf{97.4} &\textbf{91.2} \\
\end{tabular}}

	\caption{Additional experiments to evaluate the different design choices for PhaseForensics.}
	\label{tab:more_exp}
	
\end{table}

\subsection{Baselines considered in this paper.} We provide brief descriptions/details for some of the methods compared in this work.

For \textbf{Xception}~\cite{Ros19}, we follow the official training code, and include horizontal flipping. This method is an important baseline that demonstrate the utilizing per-frame features from a feature extractor followed by a classifier.

\textbf{Multi-Attention}\footnote{https://github.com/yoctta/
multiple-attention}~\cite{zhao2021multi} is a frame based method which uses a novel multi-attention mechanism combined with a local texture enhancement model. The authors argue this architecture more easily allows for the network to identify localized abnormalities in the video texture.

\textbf{CNN-GRU}~\cite{Sab19}, is a baseline to demonstrate the utilizing per-frame features from a feature extractor followed sequence modeling. We train with DenseNet-161 architecture.

Results for \textbf{PatchForensics}~\cite{Cha20}, \textbf{DSP-FWA}~\cite{Li18}, \textbf{Face X-ray}~\cite{Li20}, and \textbf{two-branch}~\cite{Mas20}, are obtained from existing benchmarks~\cite{Mas20,Hal21}, that follow similar testing protocols.

\textbf{LipForensics}\footnote{https://github.com/ahaliassos/LipForensics}~\cite{Hal21} is one of our important baselines, with a lip-based DF classifier, that learns high-level temporal semantics of the lip region.
We use the code from~\cite{Mar20}, with the last layer modified to a 1-class classifier, and use author-released weighted (trained on FaceForensics++~\cite{Ros19}) to generate the results in this paper.

For \textbf{FTCN}\footnote{https://github.com/yinglinzheng/FTCN}~\cite{Zhe21}, we utilize the evaluation code released by the authors, along with the trained weights.

\end{document}